\documentclass[10pt,twocolumn,letterpaper]{article}

\usepackage{wacv}
\usepackage{times}
\usepackage{epsfig}
\usepackage{graphicx}
\usepackage{amsmath}
\usepackage{amssymb}
\usepackage{booktabs}

\usepackage{algorithmic}
\usepackage{multirow}
\usepackage{amssymb}
\usepackage{booktabs} 
\usepackage[linesnumbered,ruled,vlined]{algorithm2e}

\newcommand*{\affaddr}[1]{#1} 
\newcommand*{\affmark}[1][*]{\textsuperscript{#1}}

\def\wacvPaperID{828} 

\wacvfinalcopy 

\ifwacvfinal
\usepackage[breaklinks=true,bookmarks=false]{hyperref}
\else
\usepackage[pagebackref=true,breaklinks=true,colorlinks,bookmarks=false]{hyperref}
\fi

\begin{document}

\title{Attend Who is Weak: Pruning-assisted Medical Image Localization under Sophisticated and Implicit Imbalances}
\author{%
Ajay Jaiswal\affmark[1], Tianlong Chen\affmark[1], Justin F. Rousseau\affmark[1], Yifan Peng\affmark[2], Ying Ding\affmark[1], Zhangyang Wang\affmark[1]\ddag\\
\vspace{-0.3cm}
\affaddr{\affmark[1]The University of Texas at Austin},
\affaddr{\affmark[2]Weill Cornell Medicine}
}

\maketitle
\thispagestyle{empty}

\begin{abstract}
   Deep neural networks (DNNs) have rapidly become a \textit{de facto} choice for  medical image understanding tasks. However, DNNs are notoriously fragile to the class imbalance in image classification. We further point out that such imbalance fragility can be amplified when it comes to more sophisticated tasks such as pathology localization, as imbalances in such problems can have highly complex and often implicit forms of presence. For example, different pathology can have different sizes or colors (w.r.t.the background), different underlying demographic distributions, and in general different difficulty levels to recognize, even in a meticulously curated balanced distribution of training data. In this paper, we propose to use pruning to automatically and adaptively identify \textit{hard-to-learn} (HTL) training samples, and improve pathology localization by attending them explicitly, during training in \textit{supervised, semi-supervised, and weakly-supervised} settings. Our main inspiration is drawn from the recent finding that deep classification models have difficult-to-memorize samples and those may be effectively exposed through network pruning \cite{hooker2019compressed} - and we extend such observation beyond classification for the first time. We also present an interesting demographic analysis which illustrates HTLs ability to capture complex demographic imbalances. Our extensive experiments on the Skin Lesion Localization task  in multiple training settings by paying additional attention to HTLs show significant improvement of localization performance by $\sim$2-3\%. 
\end{abstract}

\section{Introduction}

In the past decade, deep learning advancements have significantly influenced numerous medical imaging applications such as automated pathology diagnosis, detection, localization, and registration \cite{li2018thoracic,zhou2019multi,Jaiswal2022RoSKDAR,peng2017negbio,motlagh2018breast,ayan2019diagnosis,Jaiswal2021SCALPS,Han2020UsingRA,jaiswal2021radbert}. The success of these applications has motivated several researchers in the community to develop large-scale public datasets that can improve task performance. Although these real-world datasets have helped to build high-quality deep learning solutions, they usually suffer from class imbalance problems that can go way beyond the common perception of different training sample numbers across categories. Indeed, \textbf{imbalance in medical imaging datasets can have highly complex and subtle forms of presence}, e.g., different pathology can have different colors and sizes, and can be difficult to recognize even in meticulously curated training data. Additionally, it can further have many implicit imbalances based on gender, race, ethnicity, and demographics of individuals and can be very difficult to account for. While there have been numerous efforts to handle imbalance during DNNs training using data-level approaches such as oversampling, undersampling, synthetic sampling \cite{Kubt1997AddressingTC,Jo2004ClassIV,Batista2004ASO,Phua2004MinorityRI,rajsolomon,Chawla2002SMOTESM}, or cost-sensitive learning-based approaches \cite{lin2017focal,zhang2010cost,zhang2012decision,wang2012cost},  these efforts primarily rely on the assumption of known class distribution and overlook the more complicated forms of imbalance. The generalization ability of DNNs can suffer significantly when data imbalance is overlooked during training, resulting in poor sensitivity towards minorities and substandard performance \cite{Weiss2004MiningWR,Holte1989ConceptLA}. Many works \cite{zhang2021understanding,arpit2017closer,liu2020early,yao2020searching,han2020sigua,Xia2021RobustEH} have recently observed that DNNs tend to prioritize learning simple patterns. More concretely, the DNN optimization is content-aware, taking advantage of patterns shared by more training examples, and therefore inclined towards memorizing the majority samples. Since minority samples are underrepresented in the training set, they tend to be \textit{poorly-memorized}, and more prone to be \textit{easily-forgotten} by the model. In the \textbf{context of image classification}, a recent empirical finding by \cite{hooker2019compressed} observed that network pruning which usually removes the smallest magnitude weight in a trained DNN, disproportionately impact various classes and samples, hurting poorly-memorized samples more. In simple words, minority samples are not ``memorized well" and suffer significantly as a consequence of network pruning.

\begin{figure*}
\centering
\includegraphics[width=15cm]{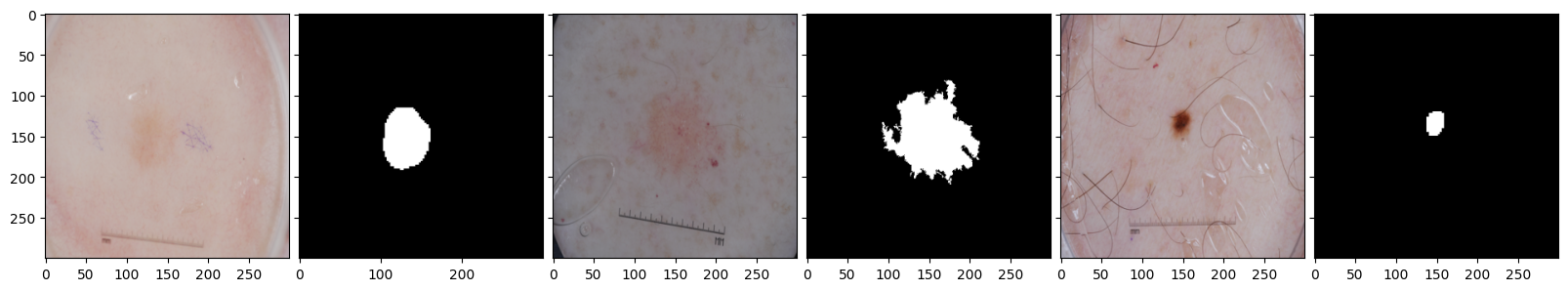}
\caption{Localization as foreground vs background classification. Spatial Imbalance can be subtle and complex (eg. indistinguishable color, irregular shape, small size etc.).}
\label{fig:foreground_vs_background}
\end{figure*}

Inspired by this observation, in this paper, we attempt to ask an interesting question: \textit{Can we identify instances which are difficult to memorize by DNNs and can be representative of complex and implicit imbalance?}. We for the \textbf{first time}, study network pruning impact on the spatial memorization/forgetting effect. We \textbf{go beyond} image classification to explore DNNs sensitivity towards \textit{instance-level spatial region imbalance}, on the real-world skin lesion localization task(S-LLT). Figure \ref{fig:foreground_vs_background} illustrate the S-LLT task as foreground vs background classification. We observed that pruning of a trained localization model has a varying impression on spatial memorization, where it significantly impacts foreground performance while having a marginal impact on the background (Figure \ref{fig:pruning_impact}).  Considering foreground as the representative of pathology and region of interest, we propose using drop-in localization (foreground) performance as a proxy to identify training instances which are poorly-memorized and can encode complex imbalance -- we call them ``\textbf{\textit{hard-to-learn}}" (HTLs) due to their high sensitivity to pruning, and show that by explicitly and adaptively paying additional attention to them during training, we can achieve notable performance gain in their localization. Interestingly, an in-depth analysis of HTLs using demographic attributes such as \textit{gender and age} reveals that \textbf{pruning impact some demographics more significantly than  others}, diligently eliciting the complex and subtle imbalances in the data, going beyond class distribution. Additionally, our work demonstrates, for the first time, that pruning can elicit the demographic bias of trained models using real-world dataset instead of curated datasets such as CIFAR and CelebA. We selected S-LLT as our evaluation task (although it can be easily adapted to any localization task) considering the technique-friendly availability of ISIC 2017, 2018 datasets \cite{codella2018skin,codella2019skin}, which will allow us to show the effectiveness of our technique in a supervised, semi-supervised, or unsupervised setting along with a detailed demographic study of HTLs. Our primary contributions can be summarized as:

\begin{figure*}
\centering
\includegraphics[width=15cm, trim={1cm 0.8cm 1cm 0.4cm}]{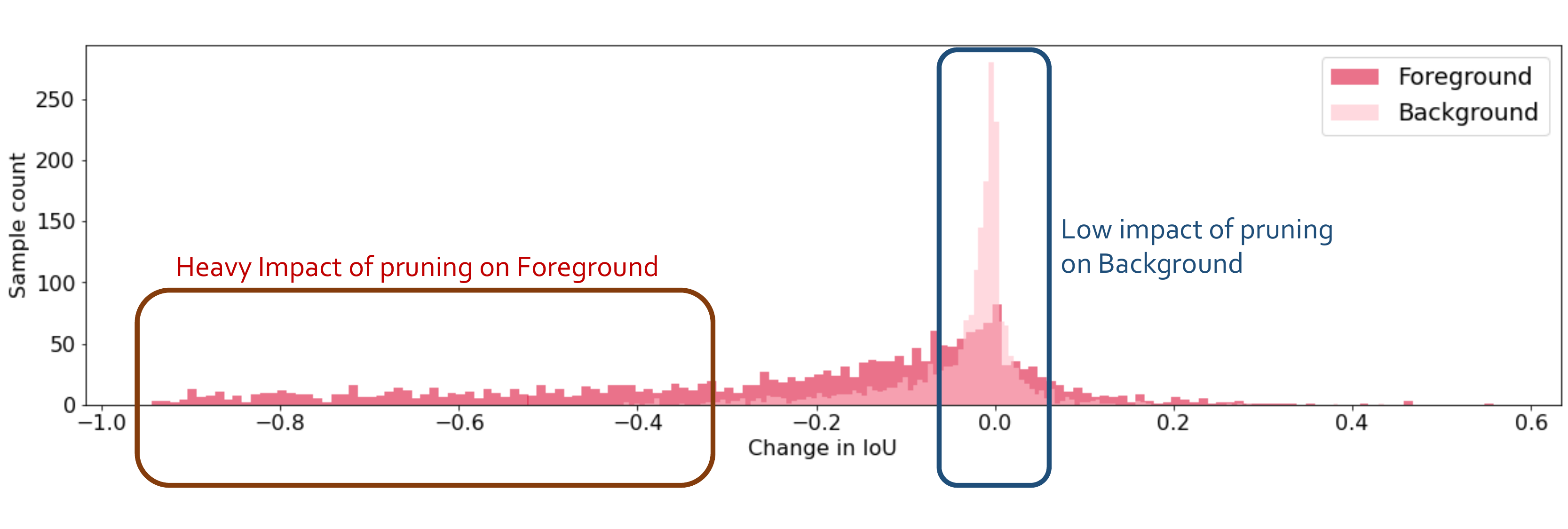}
\caption{Impact of magnitude-based pruning (70\%) of a trained U-Net based localization network on the IoU of foreground and background classes in the supervised setting. Pruning heavily impact the foreground IoU compared to background IoU.}
\label{fig:pruning_impact}
\end{figure*}

\begin{figure*}
\centering
\includegraphics[width=\linewidth]{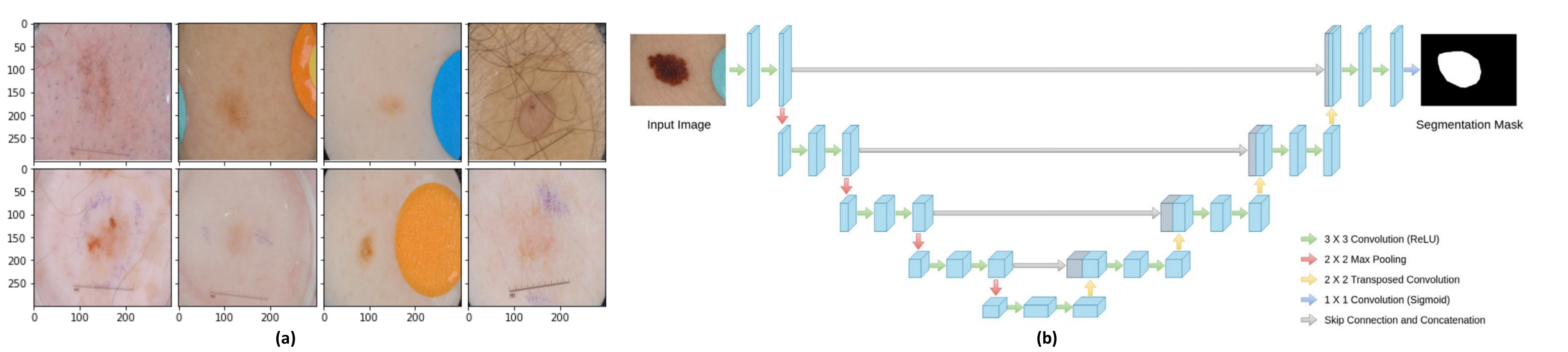}
\caption{\textbf{(a)} Randomly sampled examples of HTLs identified by pruning U-Net based localization network by 70\% in supervised setting. \textbf{(b)} Classical U-Net architecture used in our experiments for our Lesion localization. Our simple model choice ensures retaining focus on highlighting the relation between network pruning and implicit complex imbalance in dataset.} \label{fig:unet}
\end{figure*}

\begin{itemize}
    \item We propose \textbf{pruning} as an indicator to expose the spatial weakness of a trained localization model and show the existence of ``\textit{hard-to-learn}" training examples. For the first time, we reveal that pruning disproportionately impacts the foreground and background classes, where the foreground performance of some training examples can drop by a much larger margin than the background, indicating their high sensitivity to pruning. 
    
    \item Tailored for the localization problem, we present \textbf{three novel HTL mining strategies} in the supervised, semi-supervised, and weakly-supervised settings using ground truth labels, pseudo-labels, and saliency maps respectively. We additionally show that by attending HTLs by fine-tuning, we can significantly improve  localization performance.
    
    \item We have conducted extensive experiments and ablation studies to understand the specialty of HTLs on the S-LLT. Additionally, we provide an interesting \textbf{demographics analysis} of HTLs and illustrate our method's ability to capture complex implicit imbalances. Moreover, our extensive experiments show a significant and consistent \textbf{performance gain} of $\sim$2-3\% IoU across different settings for S-LLT.
\end{itemize}
\section{Methodology}
\subsection{Network pruning}
The fundamental hypothesis behind the NN pruning is that DNNs are overparameterized, and a comparatively smaller network (sparse network) can be used to achieve a similar level of performance. Provided a dataset $D = \{(x_i, y_i)\}_{i=1}^n$, and a preferred sparsity level $\kappa$ (i.e, number of non-zero weights), NN pruning can be written as a constrained optimization problem:
\begin{equation}
    \min_{w} L(\textbf{w}; D) = \min_{w} \frac{1}{n}\sum_{i=1}^{n} l(\textbf{w}; (x_{i}, y_{i})),   
\end{equation}
\begin{equation}
    s.t. \  \textbf{w}\in \mathbb{R}^m, ||w||_0 \leq \kappa 
\end{equation}

where, $l(\cdot)$ is a standard loss function, $\textbf{w}$ is a set of parameters of NN, $m$ is the total number of parameters, and $||\cdot||_0$ is $L_0$ norm. The traditional approach to minimize the above equation is by adding sparsity enforcing penalty terms or saliency-based methods. Saliency-based methods solve the above equation by removing redundant parameters in the NN using a good criterion. Popular criterion includes magnitude-based weight pruning (i.e., weight below a certain threshold is redundant) \cite{han2015learning,guo2016dynamic}, or hessian of loss wrt. weights (i.e., the higher the value of hessian, the higher the parameter importance) \cite{LeCun1989OptimalBD,Hassibi1993OptimalBS}. 

In this work, we have used magnitude-based unstructured weight pruning due to its simplicity and keep the focus on the ``\textit{forgetting}" behavior of sparse NNs \cite{hooker2019compressed}. To avoid possible confusion, we do \textbf{NOT} use pruning for any model efficiency purpose. In our framework, pruning would be better described as “selective brain damage”. It is mainly used for effectively spotting HTLs not yet well memorized and learned by the current model.

\begin{figure*}
\centering
\includegraphics[width=\linewidth, trim={0cm 1cm 1cm 1cm}]{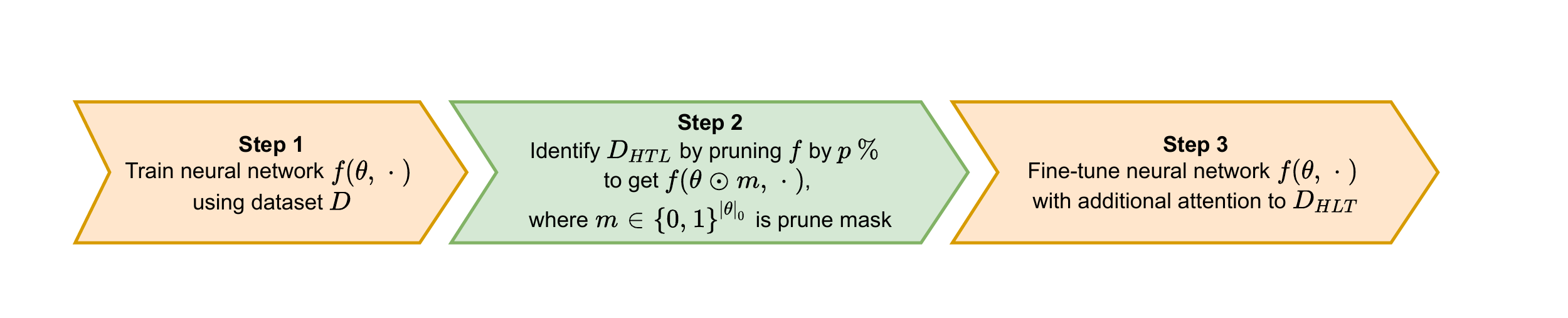}
\caption{An overview of our Pruning Assisted NN training paradigm for localization.} \label{fig:overview}
\end{figure*}

\begin{algorithm*}
\DontPrintSemicolon
  \textbf {Input:} Data $D = \{x_i\}_{i = 1}^{n}$, Localization labels $Y = \{y_i\}_{i = 1}^{n}$,  Model = $f(\theta_{init},\cdot)$, Pruning algorithm = $P$, $thresold_{HTL} = \tau$\\
  \textbf Learn $f(\theta_{intermediate},\cdot)$ by minimizing $\sum_{i=1}^{n} L_{localization}(\theta_{init}, D, Y)$\\
  \textbf Prune $f(\theta_{intermediate},\cdot)$ by $p\%$ using $P$  to get $f(\theta_{intermediate} \odot m,\cdot)$ where $m \in \{0,1\}^{\| \theta\|_{0}}$ is the prune mask.\\
  \textbf Identify $\tilde{D}_{HTL} = \{\{\tilde{x}_i, \tilde{y}_i \} $ st. $ IoU(f(\theta_{intermediate} \odot m, \tilde{x_i})) - IoU(f(\theta_{intermediate}, \tilde{x_i})) > \tau\}$ \\
  \textbf Fine-tune $f(\theta_{intermediate},\cdot) \to f(\theta_{final},\cdot)$ by minimizing $\sum_{i=1}^{\|\tilde{D}_{HTL}\|} L_{localization}(\theta_{init}, \tilde{D}_{HTL}, Y_{HTL})$\\
  \textbf Return $f(\theta_{final},\cdot)$
\caption{Supervised pruning assisted localization}
\label{alg:supervised}
\end{algorithm*}

\begin{algorithm*}
\DontPrintSemicolon
  \textbf {Input:} Data $D = \{x_i\}_{i = 1}^{n}$, Localization labels $Y = \{y_i\}_{i = 1}^{n}$, Unlabelled Data $\tilde{D} = \{\tilde{x}_i\}_{i = 1}^{k}$, Model = $f(\theta_{init},\cdot)$, Pruning algorithm = $P$, $thresold_{HTL} = \tau$\\
  \textbf Learn $f(\theta_{intermediate},\cdot)$ by minimizing $\sum_{i=1}^{n} L_{localization}(\theta_{init}, D, Y)$\\
  \textbf Generate localization pseudo-labels $\tilde{Y} = f_{\theta_{intermediate}}(\tilde{x_i})$ for $i = 1,2, ..., k$\\
  \textbf Prune $f(\theta_{intermediate},\cdot)$ by $p\%$ using $P$  to get $f(\theta_{intermediate} \odot m,\cdot)$ where $m \in \{0,1\}^{\| \theta\|_{0}}$ is the prune mask.\\
  \textbf Identify $\tilde{D}_{HTL} = \{\{\tilde{x}_i, \tilde{y}_i \} $ st. $  BB(f(\theta_{intermediate} \odot m, \tilde{x_i})) - BB(f(\theta_{intermediate}, \tilde{x_i})) > \tau\}$ \\
  \textbf Fine-tune $f(\theta_{intermediate},\cdot) \to f(\theta_{final},\cdot)$ by minimizing $\sum_{i=1}^{\|\tilde{D}_{HTL}\|} L_{localization}(\theta_{init}, \tilde{D}_{HTL}, \tilde{Y}_{HTL})$\\
  \textbf Return $f(\theta_{final},\cdot)$
\caption{Semi-supervised pruning assisted localization}
\label{alg:semi-supervised}
\end{algorithm*}

\begin{algorithm*}
\DontPrintSemicolon
  \textbf {Input:} Data $D = \{x_i\}_{i = 1}^{n}$, Classification labels $Y = \{y_i\}_{i = 1}^{n}$, Heatmap Generator = $GradCAM_{L}(\cdot)$, Model = $f(\theta_{init},\cdot)$, Pruning algorithm = $P$, $thresold_{HTL} = \tau$\\
  \textbf Learn $f(\theta_{intermediate},\cdot)$ by minimizing $\sum_{i=1}^{n} L_{classification}(\theta_{init}, D, Y)$\\
  \textbf Generate heatmap as $Y_{CAM} = GradCAM_{L}(f(\theta_{intermediate}, x_i))$ for $i = 1,2, ..., n$, where L represents tensor output of L-th layer of $f(\theta_{intermediate},\cdot)$. Note that heatmap is an indicator of localization.\\
  \textbf Prune $f(\theta_{intermediate},\cdot)$ by $p\%$ using $P$  to get $f(\theta_{intermediate} \odot m,\cdot)$ where $m \in \{0,1\}^{\| \theta\|_{0}}$ is the prune mask.\\
  \textbf Generate heatmap $\tilde{Y}_{CAM} = GradCAM_{L}(f(\theta_{intermediate} \odot m, x_i))$ for $i = 1,2, ..., n$\\
  \textbf Identify $\tilde{D}_{HTL} = \{\{\tilde{x}_i, \tilde{y}_i \}$ st. $ BB(Y_{CAM}(x_i)) - BB(\tilde{Y}_{CAM}(x_i)) > \tau\}$ \\
  \textbf Fine-tune $f(\theta_{intermediate},\cdot) \to f(\theta_{final},\cdot)$ by minimizing $\sum_{i=1}^{\|\tilde{D}_{HTL}\|} L_{classification}(\theta_{init}, \tilde{D}_{HTL}, Y_{HTL})$\\
  \textbf Return $f(\theta_{final},\cdot)$
\caption{Weakly-supervised pruning assisted localization}
\label{alg:weakly-supervised}
\end{algorithm*}

\subsection{Hard-to-Learn (HTL) Instances}
DNNs can be compressed to significantly huge levels at startlingly little loss of test accuracy using various pruning methods \cite{LeCun1989OptimalBD,wang2020picking,jaiswal2021spending,frankle2018lottery,tanaka2020pruning,lee2018snip,jaiswal2022training}. Recently, some works have identified the deeper connection of pruning with generalization/memorization, beyond considering it just as an ad-hoc compression tool \cite{jiang2021self,zhang2021efficient}. The most relevant work by \cite{hooker2019compressed} used pruning as a mean to expose the weakness of a trained model in generalization. More specifically, \cite{hooker2019compressed} identified that pruning a trained image classifier, produces a non-uniform impact on long-tail less frequent instances.
In this paper, we study this observation for the \textbf{first time}  to identify ``\textit{easily forgotten}" training examples for pathology localization in supervised, semi-supervised, and weakly-supervised settings. Using S-LLT as our experimental task, we identified that pruning disproportionately impacts the foreground and background class, hurting the foreground significantly (Figure \ref{fig:pruning_impact}). We observed that foreground performance of some training examples drops by a large margin, indicating their high sensitivity to pruning. We term these most impacted images as ``\textit{\textbf{hard-to-learn}}" (HTLs). 

We would like to highlight that our idea of HTLs is a \textbf{bold attempt} to explore beyond the class-wise label imbalance. Even artificially class-balanced datasets such as CIFAR-10/100 and ImageNet. have many hidden inherent forms of imbalance such as class-level difficulty variations or instance-level feature distribution, which reflect in the performance of trained DNNs. Since HTLs are label-agnostic and completely rely on DNN's memorization ability and learning patterns, it is applicable to various more complicated
forms of imbalance in real data, such as complex attribute imbalances \cite{Sarafianos2018DeepIA} and demographic imbalances \cite{Varoquaux2022, larrazabal2020gender}. Our in-depth analysis for S-LLT using ISIC-2017 validates the ability of HTLs to capture implicit demographic (gender and age) imbalances in the real-world dataset, which provide an opportunity to explicitly pay attention to them during training. Figure \ref{fig:unet}(a) presents some sampled examples of HTLs identified by pruning U-Net based localization network by 70\% in supervised setting. 

\subsection{Mining HTLs}
\label{sec:hlt_mining}
Medical imaging real-world datasets exhibit subtle forms of imbalances where various feature attributes have very different frequencies (eg. pathology color, size, and shape) and instance-level difficulty variations. Broadly speaking, such imbalances are not
only limited to the standard majority versus minority class but extend to implicit forms based on gender, race, ethnicity, as well as demographics of individuals and can be very
difficult to account for. In this section, we present \textbf{three different HTLs} (inherently capturing subtle imbalances in a label-agnostic way) \textbf{mining strategies} for localization in supervised, semi-supervised, and weakly-supervised settings.

\subsubsection{Supervised Setting:} Our supervised setting considers the availability of segmentation masks ($\{y_i\}_{i = 1}^{n}$) corresponding to every training image ($\{x_i\}_{i = 1}^{n}$). We first train a U-Net model (Figure \ref{fig:unet}) using the supervised cross-entropy loss ($L_{localization}$) to fit on our labeled training data. In order to identify HTLs ($\tilde{D}_{HTL}$) from the training data, we prune the trained network using a pruning algorithm $P$ by $p\%$ and look for instances which are highly sensitive to pruning (i.e., observed significant drop in foreground IoU performance). The complete supervised pruning assisted localization process is summarized in Algorithm \ref{alg:supervised}. The final network is generated by fine-tuning with additional attention to spotted HTLs.

\subsubsection{Semi-supervised Setting:} Our semi-supervised setting considers the availability of segmentation masks ($\{y_i\}_{i = 1}^{n}$) corresponding to $n$ input training images ($\{x_i\}_{i = 1}^{n}$). Additionally, it make use of $k$ pathology images ($\{\tilde{x}_i\}_{i = 1}^{k}$) for which no segmentation mask is available. We first train a U-Net model (Figure \ref{fig:unet}) using the supervised cross-entropy loss ($L_{localization}$) to fit on our labelled training data ($D$). Next, we generate pseudo-labels for $k$ unlabelled pathology images. In order to identify HTLs ($\tilde{D}_{HTL}$) from the training data, we prune the trained network using a pruning algorithm $P$ by $p\%$ and look for pseudo-labels instances which are highly sensitive to pruning (i.e., observed significant change in bounding box generated using \cite{Jaiswal2021SCALPS}). The complete semi-supervised pruning-assisted localization process is summarized in Algorithm \ref{alg:semi-supervised}. The final network is generated by fine-tuning with additional attention to  spotted HTLs.

\subsubsection{Weakly-supervised Setting:} In our weakly-supervised setting, we do not use any segmentation label corresponding to the input training dataset. Instead, we propose to use high-level classification labels to train our U-Net backbone using an additional MLP layer and supervised classification loss ($L_{classification}$). We have used ISIC-2018 dataset, which provides 10,015 images without segmentation masks divided into eight different clinical scenarios and evaluated performance on the ISIC-2017 test set with segmentation labels. In this setting, we pass the feature tensor from the last convolutional layer of the U-Net model trained with classification loss to GradCAM$++$\cite{Chattopadhyay2018GradCAMGG} and generate a bounding box using \cite{Jaiswal2021SCALPS}. To identify HTLs, we compare the bounding boxes generated for input images before and after pruning the backbone network by $p\%$ using the pruning algorithm $P$. We summarize the complete weakly-supervised pruning-assisted localization process in Algorithm \ref{alg:weakly-supervised}. The final network is generated by fine-tuning with additional attention to  HTLs.

\subsection{Unified Pipeline} Our completed pruning-assisted localization pipeline is presented in Figure \ref{fig:overview}. Given a neural network $f(\theta, \cdot)$, we first train $f$ using the training dataset $D$. Next, we identify HTLs using methods proposed in section \ref{sec:hlt_mining} which highlights the weakness of our trained network $f(\theta, \cdot)$. Finally, we fine-tune our network $f$ by paying additional attention to HTLs using weighted cross-entropy loss as:
\begin{equation}
    l_n = - \sum_{i=1}^C w_c \times \log \frac{\exp(\tilde{x}_{n,c})}{\exp (\sum_{i=1}^C \tilde{x}_{n,i})} \times \tilde{y}_{n,c}
\end{equation}

where $\tilde{x} \in \tilde{D}_{HTL}$, $\tilde{y}$ is the target, $w$ is the weight of class $c$, $C$ is the number of classes (i.e, foreground, background), and n is n-th training example. Our extensive experimental analysis indicates that the fine-tuned network achieve significantly high foreground performance gain on the ISIC 2017 test set across all three training paradigms.

\section{Experimental Settings}
\paragraph{Dataset Details:} Our experiments used skin lesion localization (S-LLT) as our evaluation task and acquired dermoscopic images from the ISIC-2017 \cite{codella2018skin} and ISIC-2018 \cite{codella2019skin} challenge. The ISIC-2017 dataset consists of 2000, 150, and 600 lesion images in JPEG format for training, validation, and test along with the corresponding expert-annotated binary segmentation mask images in PNG format. It additionally provides demographic metadata entries of age and sex for the patients which we have used to validate our method's ability to elicit complex and sophisticated demographic imbalances. For our semi- and weakly- supervised task we have used ISIC-2018 dataset, which provides 10,015 images without segmentation masks divided into eight different clinical scenarios. Although our method can be adapted to any task, ISIC datasets provide metadata information (demographic details such as gender and age) along with high-quality segmentation annotations as well as unannotated images with classification labels. This facilitates a unique opportunity to effectively evaluate the benefits of our proposed method (Section \ref{sec:hlt_mining}) in multiple training settings (i.e., supervised, semi-supervised, and weakly-supervised) along with demographic analysis.


\begin{table*}
\small
\centering
\begin{tabular}{ccccccccc} 
 \toprule
 \small
 \multirow{2}{*}{}  & \multicolumn{4}{c}{\textbf{Supervised}} & \multicolumn{4}{c}{\textbf{Semi-supervised}} \\ 
 \cmidrule(rr){0-0}
 \cmidrule(rr){2-5}
 \cmidrule(rr){6-9}
 \textbf{Prune Ratio} & 25\% & 50\% & 75\% & 99\% & 25\% & 50\% & 75\% & 99\% \\
 \midrule
 Foreground & 12.41\% ($\downarrow$) & 26.64\% ($\downarrow$) & 42.19\% ($\downarrow$) & 86.15\% ($\downarrow$) & 9.72\% ($\downarrow$) & 15.33\% ($\downarrow$) & 43.01\% ($\downarrow$) & 81.98\% ($\downarrow$)\\
 Background & 0.95\% ($\downarrow$) & 6.03\% ($\downarrow$) & 17.11\% ($\downarrow$) & 48.30\% ($\downarrow$) & 0.26\% ($\downarrow$) & 4.87\% ($\downarrow$) & 11.32\% ($\downarrow$) & 47.71\% ($\downarrow$) \\
\bottomrule
\end{tabular}
\vspace{0.1cm}
\caption{Percentage drop in IoU of training samples in ISIC-2017 dataset when the trained network is pruned by $\mathbf{p}\%$ using unstructured magnitude-based pruning in supervised and semi-supervised settings.  Disproportionate impact of pruning can be clearly observed for the background and foreground classes.}
\label{table:pruning_impact}
\end{table*}

\begin{figure*}
\centering
\includegraphics[width=13cm]{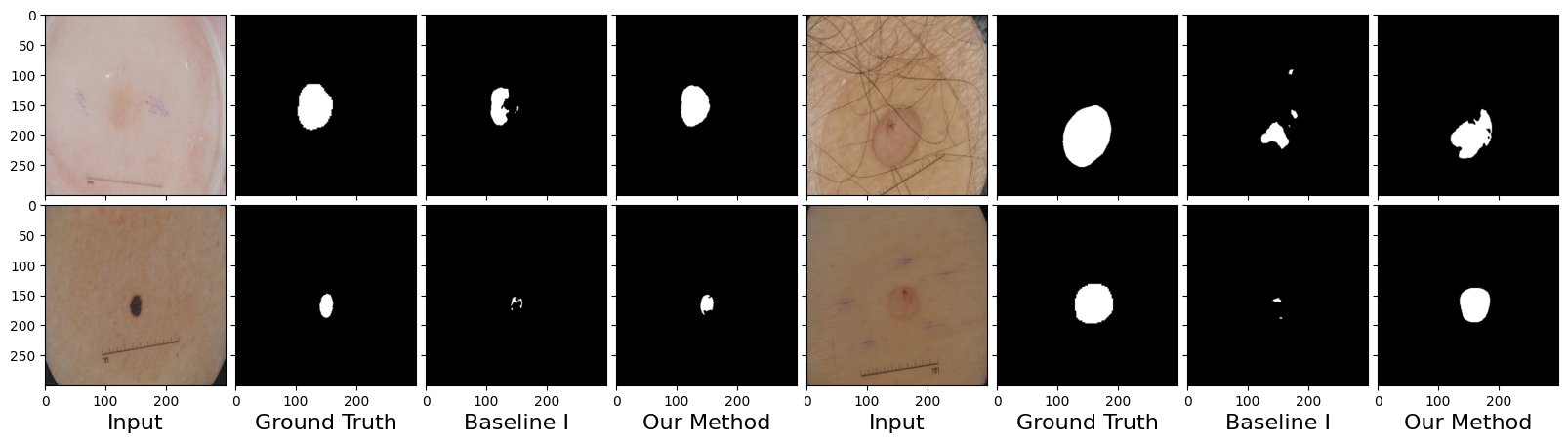}
\caption{Examples of the visualization of segmentation mask generated by Baseline 1 and our method on HTL samples. With additional attention to HTL samples in fine-tuning, the model is able to generate better segmentation masks for HTLs improving overall performance.} 
\label{fig:hlt_results}
\vspace{-0.3cm}
\end{figure*}

\vspace{-0.3cm}
\paragraph{Training and Fine-tuning Details:} In our experiments, all models are trained using similar settings and seed values (10, 20, 30). We have used an SGD optimizer with a momentum of 0.9 and weight decay of $2e^{-4}$. The initial learning rate is set to 0.1, and the networks are trained for 100 epochs with a batch size of 64. The learning rate decays by a factor of 10 at the [20, 50, 80]th epoch during the training.  We have used standard augmentation techniques to flip, rotate, and mirror the images during training. For pruning, we have used a global unstructured-magnitude based pruning and prune ratio of 70\% across all experiments. During the fine-tuning stage, we started with a smaller learning rate of 0.01 and retrained the network with HTLs for 20 epochs with a decay at the 15th epoch. All our models have been trained using 4 Quadro RTX 5000 GPUs and we have evaluated our models using foreground and background IoU scores against different baselines.

\vspace{-0.3cm}
\paragraph{Additional Implementation Utility Details :} In all our experiments (except ablation), we have used unstructured magnitude-based pruning, where we find and remove the least salient connections (weight magnitude) in the model wherever they are. For bounding box generation in Algorithm \ref{alg:semi-supervised} and \ref{alg:weakly-supervised}, we pass the feature tensor of the last convolution layer to GradCAM++\cite{Chattopadhyay2018GradCAMGG} to extract the heatmaps. We further scale the heatmap intensity to the range [0-255] and use an ad-hoc threshold (pixel value = 180) to binarize the heatmap. In last, we followed the pseudocode proposed in \cite{Jaiswal2021SCALPS} to create the bounding box. We have used two popular evaluation metrics (IoU and DICE) to compare our method performance against different baselines. 


\vspace{-0.3cm}
\paragraph{Baseline Comparison:}  In our experiments, the first baseline is a U-Net architecture trained to perform the S-LLT. We have adapted the original vanilla U-Net (\textbf{Baseline 1}) version proposed in \cite{ronneberger2015u} to avoid any design overhead and highlight the importance of HTLs. Our second baseline is the top-performing architecture from ISIC-2017 challenge leaderboard (\textbf{Baseline 2}). Our third baseline uses focal loss \cite{lin2017focal}, which has been one default choice to handle imbalance (\textbf{Baseline 3}). In our fourth baseline,  we randomly sample exactly the same number of instances (not specifically picked HTLs) and fine-tune our network similar to our proposed method, to validate the significance of identifying and using HTLs (\textbf{Baseline 4}). Next, in our fifth baseline, we randomly sampled exactly the same number of instances following the class distribution in S-LLT dataset where minority classes are sampled with higher probability to fine-tune our network (\textbf{Baseline 5}). Lastly, our final baseline randomly sampled exactly the same number of instances following the demographics distribution (gender) to fine-tune our network (\textbf{Baseline 6}). The performance comparison of all baselines compared to our HTL-based fine-tuning is reported in Table \ref{table:main_results}, which clearly unveil the effectiveness of our method. Note that the \textit{main goal of our work} is to elicit the effectiveness of network pruning in identifying complex implicit imbalances in medical datasets, and propose a simple and unified approach to identify data samples suffering from imbalance during training, rather than proposing a task-specific novel class imbalance algorithm.

\begin{table*}
\scriptsize
\centering
\begin{tabular}{ccccccccccccc} 
 \toprule
 \small
 \multirow{2}{*}{\textbf{Method}}  & \multicolumn{4}{c}{\textbf{Supervised}} & \multicolumn{4}{c}{\textbf{Semi-supervised}} & \multicolumn{4}{c}{\textbf{Weakly-supervised}}\\ 
\cmidrule(rr){2-5}
\cmidrule(rr){6-9}
\cmidrule(rr){10-13}
 & B(IOU) & F(IoU) & Mean(IoU) & DICE & B(IoU) & F(IoU) & Mean(IoU) & DICE & B(IoU) & F(IoU) & Mean(IoU) & DICE\\
 \midrule
 
 Baseline 1 & 0.894 & 0.654 & 0.774 & 0.855 & 0.901 & 0.661 & 0.781 & 0.865 & 0.622 & 0.420 & 0.521 & 0.637\\
 Baseline 2 & 0.891 & 0.637 & 0.764 & 0.849 & 0.876 & 0.642 & 0.759 & 0.813 & 0.568 & 0.390 & 0.479 & 0.596\\
 Baseline 3 & 0.892 & 0.663 & 0.775 & 0.863 & 0.906 & 0.669 & 0.788 & 0.872 & 0.614 & 0.411 & 0.513 & 0.644\\
 Baseline 4 & 0.883 & 0.660 & 0.772 & 0.852 & 0.894 & 0.662 & 0.778 & 0.866 & 0.609 & 0.407 & 0.508 & 0.641\\
 Baseline 5 & 0.878 & 0.659 & 0.769 & 0.844 & 0.900 & 0.647 & 0.774 & 0.859 & 0.610 & 0.399 & 0.504 & 0.635\\
 Baseline 6 & 0.880 & 0.665 & 0.773 & 0.859 & 0.891 & 0.671 & 0.781 & 0.871 & 0.619 & 0.413 & 0.516 & 0.640\\
 \midrule
 \textbf{Our Method} & 0.890 & 0.681 & 0.785 & 0.871 & 0.904 & 0.695 & 0.800 & 0.884 & 0.643 & 0.438 & 0.542 & 0.657\\
 \textbf{Std} & $\pm$0.002 & $\pm$0.001& $\pm$0.001& $\pm$0.003& $\pm$0.001& $\pm$0.001& $\pm$0.001& $\pm$0.002& $\pm$0.006& $\pm$0.008& $\pm$0.007& $\pm$0.005\\
\bottomrule
\end{tabular}
\vspace{0.1cm}
\caption{Performance comparison of our method against different baselines using unstructured magnitude-based pruning with prune ratio of 70\% on ISIC-2017 test set. Our method significantly improves the Foreground performance compared to all baselines.}
\label{table:main_results}

\end{table*}

\begin{figure*}
    \centering
    \includegraphics[width=14cm]{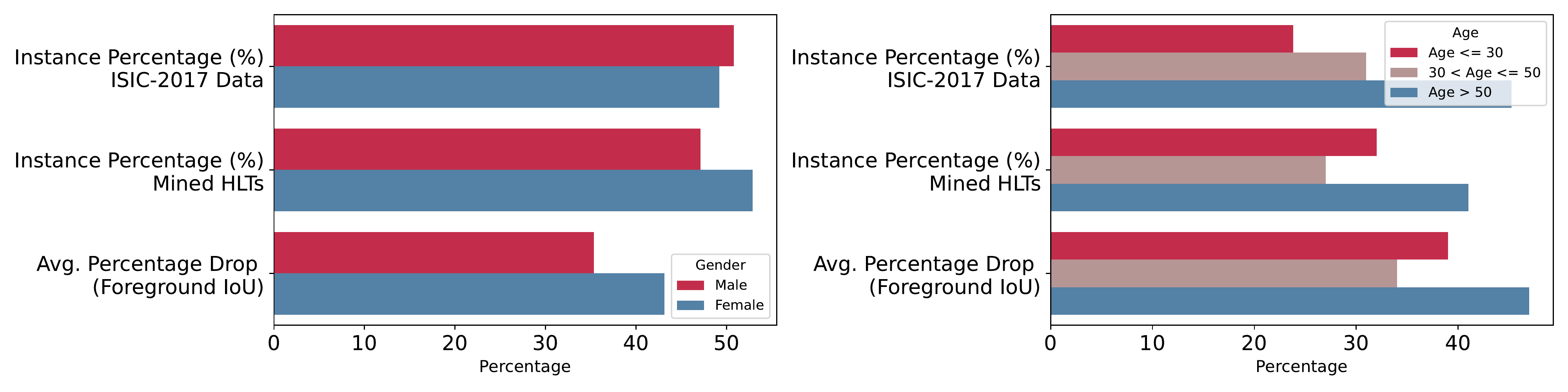}
    \caption{(a) Gender distribution, (b) Age distribution of instances in ISIC-2017, and mined HTLs using unstructured pruning (70\%) of trained U-Net in supervised setting. Clearly, pruning differently impacts different age and gender groups' foreground performance.}
    \label{fig:demographics}
    \vspace{-0.3cm}
\end{figure*}

\section{Main Results and Discussion}
In this section, we present a comprehensive analysis of the impact of pruning in eliciting the weakness of trained localization models, and its high sensitivity towards the foreground performance compared to the background. Additionally, we provided a performance comparison of our proposed pruning-assisted localization algorithms \ref{alg:supervised},\ref{alg:semi-supervised}, and \ref{alg:weakly-supervised} against several baselines. Finally, we provide an interesting observation of demographic bias captured by our pruning-based HTLs, where some demographics have a higher impact of pruning compared to others. Last, we have conducted an ablation study to illustrate that our observations are agnostic to various pruning methods, and HTLs identified by any pruning methods have similar benefits.

\vspace{-0.3cm}
\paragraph{\textit{How does pruning impact foreground vs background?}} We find that pruning consistently amplifies the disparate treatment of foreground performance for all levels of compression we consider. Figure \ref{fig:pruning_impact} illustrate when a U-Net based trained localization network is pruned by 70\% using magnitude-based unstructured pruning. We observe that nearly $\sim$18\% instances lose their foreground performance (IoU) significantly by $>=$40\%, compared to $\sim$1\% losing their background performance (IoU) by  $>=$40\%. Additionally, Table \ref{table:pruning_impact}, presents a detailed analysis of the disproportionate impact of pruning with varying thresholds $p \in \{25\%, 50\%, 75\%, 99\%\}$ on the foreground and background IoU in supervised and semi-supervised settings. It can be observed that across all the pruning thresholds, the foreground suffers more than the background in both training settings. Note that at very high sparsity (eg. 99\%), the network performance becomes significantly low, and even the background performance drop significantly by $>$ 47\%, making it unsuitable for HTL mining. Based on our experiments, we recommend 60-80\% pruning ratio for unstructured magnitude-based pruning. 

\begin{table*}
\vspace{-0.4cm}
\centering
\small
\begin{tabular}{llllll}
\toprule
Method &  Full Network & 20\% & 70\% & 95\%\\
\midrule
Random Pruning & 0.654 & 0.431 ($\downarrow$0.223) & 0.218 ($\downarrow$0.436) & 0.022 ($\downarrow$0.632)\\
Unstructured MB-Pruning & 0.654 & 0.580 ($\downarrow$0.074) & 0.409 ($\downarrow$0.245) & 0.296 ($\downarrow$0.358)\\
Structured MB-Pruning & 0.654 & 0.533 ($\downarrow$0.121) & 0.387 ($\downarrow$0.267) & 0.215 ($\downarrow$0.439) \\
\bottomrule
\end{tabular}
\vspace{0.1cm}
\caption{Impact of various pruning methods with varying sparsity on foreground performance in supervised setting. }
\label{table:prune_percentage}
\vspace{-0.4cm}
\end{table*}

\vspace{-0.3cm}
\paragraph{\textit{How does HTL-aware training benefit overall performance?}}  In our work, we hypothesized HTLs to be representative of prevalent complex and implicit imbalance in the medical imaging dataset. In order to justify the benefits of workload to capture HTLs, it is important to illustrate how they can help in improving localization performance. Table \ref{table:main_results} presents the comparison of our method (mean across 3 independent runs with seed 10, 20, 30) against three aforementioned baselines for supervised, semi-supervised, and weakly-supervised settings. It can be clearly observed that our pruning-assisted algorithms \ref{alg:supervised}, \ref{alg:semi-supervised}, and \ref{alg:weakly-supervised}, provide consistent and significant performance gain for two popular localization metrics IoU and DICE across all three training settings. More precisely, our proposed method achieve significant gain (IoU) of $+2.7\%, +3.4\%, $ and $+2.1\%$ over baseline 1 in supervised, semi-supervised, and weakly-supervised settings for desired foreground class. Our method performance is significantly high when compared with ISIC leaderboard performance (Baseline 2). Focal loss proposed in \cite{lin2017focal} has been a very popular choice to handle imbalance but it is limited by the requirement of class distribution and the inability to capture subtle implicit imbalances. Compared to focal loss based baseline 3, our method achieves $+1.0\%, +1.2\%, and +2.9\%$ better performance for three training settings.

To confirm that HTLs are special instances, our Baseline 4 randomly samples exactly same number of training instances in the dataset followed by fine-tuning using the same training protocols as HTLs. Table \ref{table:main_results} elucidate the importance of HTLs when compared to baseline 4. Next, to our surprise Baseline 5, which samples fine-tuning instances following the class distribution, performs significantly worse than Baseline 4, which randomly samples HTLs without following any class distribution. An in-depth analysis reveals that although minority classes \textit{dermatofibroma} and \textit{vascular lesion} corresponds to only $\sim0.9\% $ and $\sim1\%$ in the dataset, they have \textbf{almost perfect performance} on the test set (98.7\%, and 98.6\% respectively). \textit{Sampling additional samples to fine-tune these classes doesn't help (possibly leads to over-fitting on minority samples)}. This clearly bolsters our motivation that the imbalances are not only limited to the standard majority versus minority class but extend to implicit and subtle forms such as feature attributes  (e.g., pathology color, size, and shape) and based on demographics (e.g., gender, race, ethnicity), and can be very difficult to account for. Lastly, the proportion of diagnoses  associated with males and females is 50.81\% and 49.19\% in S-LLT dataset. Our last Baseline 6, which randomly samples following the gender distribution, has marginal benefit over Baseline 4 ($+0.5\%$ foreground performance), which further suggests the complexity of imbalance. Finally, Figure \ref{fig:hlt_results} illustrates a visualization of the segmentation mask generated by Baseline 1 and our method on some randomly selected samples from HTL mining. With additional attention to HTL samples in fine-tuning, the UNet-based localization model can generate better segmentation masks for HTL instances, improving the overall performance.

\vspace{-0.2cm}
\begin{figure}[h]
    \centering
    \includegraphics[width=\linewidth]{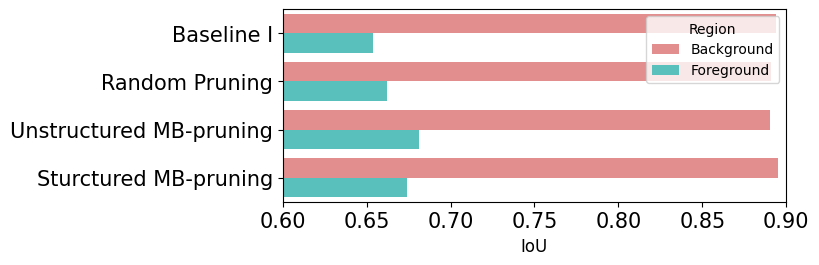}
    \caption{Localization performance of different pruning methods in supervised setting (HTLs mined with prune ratio 70\%).}
    \label{fig:prune_ablation}
    \vspace{-0.4cm}
\end{figure}

\vspace{-0.3cm}
\paragraph{\textit{How do HTLs capture subtle demographic imbalance?}} HTLs are our bold attempt toward exploring complex and implicit imbalances which go beyond the class distribution in localization settings. In this section, we present an interesting in-depth demographic analysis of HTLs and validate that our pruning-assisted HTL mining indeed captures the demographic bias of the trained U-Net localization network. Figure \ref{fig:demographics} illustrates the gender and age group distribution of data points in ISIC-2017 dataset and mined HTLs.  It can be clearly observed that although ISIC-2017 is fairly balanced from the gender perspective (50.82\% and 49.18\% for males and females), mined HTLs subset has a skewed gender distribution favoring females by $\sim$ 4\%. Moreover, the average \textbf{drop} in foreground performance for females is $>$8\% compared to men, which is an indicator of the model \textbf{favoritism} towards learning instances belonging to men patients. In addition, from the age group perspective, it can be observed that patients belonging to the age group within 30-40 years constitute the second largest proportion in data ($\sim$30.95\%), but they are least impacted by pruning and share only $\sim$26.52\% in HTLs subset and have a minimal drop in foreground IoU performance. However, interestingly, patients belonging to the age group $<=$30 years which have the least share in the data ($\sim$23.79\%), are heavily impacted by pruning and contribute $\sim$32.11\% in the HTLs subset which again points out the model's difficulty to learn patients belonging to this age group. This analysis is a \textbf{strong} indicator of  pruning ability to uncover demographic bias in medical imaging datasets in localization tasks.

\vspace{-0.3cm}
\paragraph{\textit{Impact of pruning algorithms on performance:}} To investigate that our observation of the disproportionate impact of pruning is agnostic to pruning methods, we carry out ablation studies on random pruning, unstructured magnitude-based pruning, and structured magnitude-based pruning. Table \ref{table:prune_percentage} illustrates the disproportionate impact of pruning methods with varying pruning ratio $p \in \{0\%, 20\%, 70\%, 95\%\}$ on foreground, with marginal impact on background IoU. In our experiments, we observed that $p$ has minimal sensitivity to performance, and we achieve approximately similar performance gain of $\sim$2.812\%$\pm$0.369 for $p \in \{50\%, 60\%, 70\%, 80\%\}$ in the supervised setting. Note that a large value of $p$ will lead the pruned network to forget a lot of information, and due to overparametrization of DNNs, a small value of $p$ will have no effect. Note that in all our experiments, we have used a pruning ratio of 70\% considering its slightly better performance.  Lastly, Figure \ref{fig:prune_ablation} illustrates the localization performance of different pruning methods in supervised settings. Clearly, it can be observed that without hurting background IoU, all the pruning methods help in improving the foreground IoU significantly. 
\section{Conclusion}
Contrary to the popular usage of pruning as an ad-hoc compression tool, in this paper, we present pruning as a technique to expose the weakness of a trained localization model and show the existence of ``textit{hard-to-learn}" training examples. We present three HTL mining strategies in  supervised, semi-supervised, and weakly-supervised settings using ground truth labels, pseudo-labels, and saliency maps. We experimentally show that by attending HTLs during fine-tuning, we can significantly improve  localization performance. Lastly, we present an interesting demographic analysis which illustrates
HTLs ability to capture complex demographic imbalances. Our future work will aim for more theoretical understanding of the HTLs and their significance. 

\section*{Acknowledgment}

This work is supported by the National Library of Medicine under Award No. 4R00LM013001 and National NSF AI Center at UT Austin.

\newpage

{\small
\bibliographystyle{ieee_fullname}
\bibliography{egbib}

\begin{thebibliography}{10}\itemsep=-1pt

\bibitem{arpit2017closer}
Devansh Arpit, Stanislaw Jastrzebski, Nicolas Ballas, David Krueger, Emmanuel
  Bengio, Maxinder~S Kanwal, Tegan Maharaj, Asja Fischer, Aaron Courville,
  Yoshua Bengio, et~al.
\newblock A closer look at memorization in deep networks.
\newblock In {\em ICML 2017}, pages 233--242. PMLR, 2017.

\bibitem{ayan2019diagnosis}
Enes Ayan and Halil~Murat {\"U}nver.
\newblock Diagnosis of pneumonia from chest x-ray images using deep learning.
\newblock In {\em 2019 Scientific Meeting on Electrical-Electronics \&
  Biomedical Engineering and Computer Science (EBBT)}, pages 1--5. Ieee, 2019.

\bibitem{Batista2004ASO}
Gustavo E. A. P.~A. Batista, Ronaldo~Cristiano Prati, and Maria~Carolina
  Monard.
\newblock A study of the behavior of several methods for balancing machine
  learning training data.
\newblock {\em SIGKDD}, 6:20--29, 2004.

\bibitem{Chattopadhyay2018GradCAMGG}
Aditya Chattopadhyay, Anirban Sarkar, Prantik Howlader, and Vineeth~N.
  Balasubramanian.
\newblock Grad-cam++: Generalized gradient-based visual explanations for deep
  convolutional networks.
\newblock {\em WACV 2018}, pages 839--847, 2018.

\bibitem{Chawla2002SMOTESM}
N. Chawla, K. Bowyer, Lawrence~O. Hall, and W.~Philip Kegelmeyer.
\newblock Smote: Synthetic minority over-sampling technique.
\newblock {\em J. Artif. Intell. Res.}, 16:321--357, 2002.

\bibitem{codella2019skin}
Noel Codella, Veronica Rotemberg, Philipp Tschandl, M~Emre Celebi, Stephen
  Dusza, et~al.
\newblock Skin lesion analysis toward melanoma detection 2018: A challenge
  hosted by the international skin imaging collaboration (isic).
\newblock {\em arXiv preprint arXiv:1902.03368}, 2019.

\bibitem{codella2018skin}
Noel~CF Codella, David Gutman, M~Emre Celebi, Brian Helba, Michael~A Marchetti,
  et~al.
\newblock Skin lesion analysis toward melanoma detection: A challenge at the
  2017 international symposium on biomedical imaging (isbi), hosted by the
  international skin imaging collaboration (isic).
\newblock In {\em 2018 IEEE 15th international symposium on biomedical imaging
  (ISBI 2018)}, pages 168--172. IEEE, 2018.

\bibitem{frankle2018lottery}
Jonathan Frankle and Michael Carbin.
\newblock The lottery ticket hypothesis: Finding sparse, trainable neural
  networks.
\newblock {\em arXiv preprint arXiv:1803.03635}, 2018.

\bibitem{guo2016dynamic}
Yiwen Guo, Anbang Yao, and Yurong Chen.
\newblock Dynamic network surgery for efficient dnns.
\newblock {\em Advances in neural information processing systems}, 29, 2016.

\bibitem{han2020sigua}
Bo Han, Gang Niu, Xingrui Yu, Quanming Yao, Miao Xu, Ivor Tsang, and Masashi
  Sugiyama.
\newblock Sigua: Forgetting may make learning with noisy labels more robust.
\newblock In {\em International Conference on Machine Learning}, pages
  4006--4016. PMLR, 2020.

\bibitem{han2015learning}
Song Han, Jeff Pool, John Tran, and William Dally.
\newblock Learning both weights and connections for efficient neural network.
\newblock {\em Advances in neural information processing systems}, 2015.

\bibitem{Han2020UsingRA}
Yan Han, Chongyan Chen, Liyan Tang, Mingquan Lin, Ajay Jaiswal, Ying Ding, and
  Yifan Peng.
\newblock Using radiomics as prior knowledge for abnormality classification and
  localization in chest x-rays.
\newblock {\em ArXiv}, abs/2011.12506, 2020.

\bibitem{Hassibi1993OptimalBS}
Babak Hassibi, David~G. Stork, and Gregory~J. Wolff.
\newblock Optimal brain surgeon and general network pruning.
\newblock {\em IEEE International Conference on Neural Networks}, 1993.

\bibitem{Holte1989ConceptLA}
Robert~C. Holte, Liane Acker, and Bruce~W. Porter.
\newblock Concept learning and the problem of small disjuncts.
\newblock In {\em IJCAI}, 1989.

\bibitem{hooker2019compressed}
Sara Hooker, Aaron Courville, Gregory Clark, Yann Dauphin, and Andrea Frome.
\newblock What do compressed deep neural networks forget?
\newblock {\em arXiv preprint arXiv:1911.05248}, 2019.

\bibitem{Jaiswal2022RoSKDAR}
Ajay Jaiswal, Kumar Ashutosh, Justin~F. Rousseau, Yifan Peng, Zhangyang Wang,
  and Ying Ding.
\newblock Ros-kd: A robust stochastic knowledge distillation approach for noisy
  medical imaging.
\newblock 2022.

\bibitem{Jaiswal2021SCALPS}
Ajay Jaiswal, Tianhao Li, Cyprian Zander, Yan Han, Justin~F. Rousseau, Yifan
  Peng, and Ying Ding.
\newblock Scalp - supervised contrastive learning for cardiopulmonary disease
  classification and localization in chest x-rays using patient metadata.
\newblock {\em 2021 IEEE International Conference on Data Mining (ICDM)}, pages
  1132--1137, 2021.

\bibitem{jaiswal2022training}
Ajay Jaiswal, Haoyu Ma, Tianlong Chen, Ying Ding, and Zhangyang Wang.
\newblock Training your sparse neural network better with any mask.
\newblock {\em arXiv preprint arXiv:2206.12755}, 2022.

\bibitem{jaiswal2021radbert}
Ajay Jaiswal, Liyan Tang, Meheli Ghosh, Justin~F Rousseau, Yifan Peng, and Ying
  Ding.
\newblock Radbert-cl: Factually-aware contrastive learning for radiology report
  classification.
\newblock In {\em Machine Learning for Health}, pages 196--208. PMLR, 2021.

\bibitem{jaiswal2021spending}
Ajay~Kumar Jaiswal, Haoyu Ma, Tianlong Chen, Ying Ding, and Zhangyang Wang.
\newblock Spending your winning lottery better after drawing it.
\newblock {\em arXiv preprint arXiv:2101.03255}, 2021.

\bibitem{jiang2021self}
Ziyu Jiang, Tianlong Chen, Bobak~J Mortazavi, and Zhangyang Wang.
\newblock Self-damaging contrastive learning.
\newblock In {\em International Conference on Machine Learning}, pages
  4927--4939. PMLR, 2021.

\bibitem{Jo2004ClassIV}
Taeho Jo and Nathalie Japkowicz.
\newblock Class imbalances versus small disjuncts.
\newblock {\em SIGKDD Explor.}, 6:40--49, 2004.

\bibitem{Kubt1997AddressingTC}
Miroslav Kub{\'a}t and Stan Matwin.
\newblock Addressing the curse of imbalanced training sets: One-sided
  selection.
\newblock In {\em ICML}, 1997.

\bibitem{larrazabal2020gender}
Agostina~J Larrazabal, Nicol{\'a}s Nieto, Victoria Peterson, Diego~H Milone,
  and Enzo Ferrante.
\newblock Gender imbalance in medical imaging datasets produces biased
  classifiers for computer-aided diagnosis.
\newblock {\em Proceedings of the National Academy of Sciences},
  117(23):12592--12594, 2020.

\bibitem{LeCun1989OptimalBD}
Yann LeCun, John~S. Denker, and Sara~A. Solla.
\newblock Optimal brain damage.
\newblock In {\em NIPS}, 1989.

\bibitem{lee2018snip}
Namhoon Lee, Thalaiyasingam Ajanthan, and Philip~HS Torr.
\newblock Snip: Single-shot network pruning based on connection sensitivity.
\newblock {\em arXiv preprint arXiv:1810.02340}, 2018.

\bibitem{li2018thoracic}
Zhe Li, Chong Wang, Mei Han, Yuan Xue, Wei Wei, Li-Jia Li, and Li Fei-Fei.
\newblock Thoracic disease identification and localization with limited
  supervision.
\newblock In {\em CVPR}, 2018.

\bibitem{lin2017focal}
Tsung-Yi Lin, Priya Goyal, Ross Girshick, Kaiming He, and Piotr Doll{\'a}r.
\newblock Focal loss for dense object detection.
\newblock In {\em Proceedings of the IEEE international conference on computer
  vision}, pages 2980--2988, 2017.

\bibitem{liu2020early}
Sheng Liu, Jonathan Niles-Weed, Narges Razavian, and Carlos Fernandez-Granda.
\newblock Early-learning regularization prevents memorization of noisy labels.
\newblock {\em Advances in neural information processing systems},
  33:20331--20342, 2020.

\bibitem{motlagh2018breast}
Mehdi~Habibzadeh Motlagh, Mahboobeh Jannesari, HamidReza Aboulkheyr, Pegah
  Khosravi, Olivier Elemento, Mehdi Totonchi, and Iman Hajirasouliha.
\newblock Breast cancer histopathological image classification: A deep learning
  approach.
\newblock {\em BioRxiv}, page 242818, 2018.

\bibitem{peng2017negbio}
Yifan Peng, Xiaosong Wang, Le Lu, Mohammadhadi Bagheri, Ronald Summers, and
  Zhiyong Lu.
\newblock Negbio: a high-performance tool for negation and uncertainty
  detection in radiology reports, 2017.

\bibitem{Phua2004MinorityRI}
Clifton Phua, Damminda Alahakoon, and Vincent Cheng-Siong Lee.
\newblock Minority report in fraud detection: classification of skewed data.
\newblock {\em SIGKDD Explor.}, 6:50--59, 2004.

\bibitem{rajsolomon}
Mayank Raj, Ajay Jaiswal, Rohit R.R, Ankita Gupta, Sudeep~Kumar Sahoo, Vertika
  Srivastava, and Yeon~Hyang Kim.
\newblock {S}olomon at {S}em{E}val-2020 task 11: Ensemble architecture for
  fine-tuned propaganda detection in news articles.
\newblock In {\em Proceedings of the Fourteenth Workshop on Semantic
  Evaluation}, pages 1802--1807, Barcelona (online), Dec. 2020. International
  Committee for Computational Linguistics.

\bibitem{ronneberger2015u}
Olaf Ronneberger, Philipp Fischer, and Thomas Brox.
\newblock U-net: Convolutional networks for biomedical image segmentation.
\newblock In {\em International Conference on Medical image computing and
  computer-assisted intervention}, pages 234--241. Springer, 2015.

\bibitem{Sarafianos2018DeepIA}
Nikolaos Sarafianos, Xiang Xu, and I. Kakadiaris.
\newblock Deep imbalanced attribute classification using visual attention
  aggregation.
\newblock {\em ArXiv}, abs/1807.03903, 2018.

\bibitem{tanaka2020pruning}
Hidenori Tanaka, Daniel Kunin, Daniel~LK Yamins, and Surya Ganguli.
\newblock Pruning neural networks without any data by iteratively conserving
  synaptic flow.
\newblock {\em preprint arXiv:2006.05467}, 2020.

\bibitem{Varoquaux2022}
Gaël Varoquaux and Veronika Cheplygina.
\newblock Machine learning for medical imaging: methodological failures and
  recommendations for the future.
\newblock {\em npj Digital Medicine}, 5(1), Apr. 2022.

\bibitem{wang2020picking}
Chaoqi Wang, Guodong Zhang, and Roger Grosse.
\newblock Picking winning tickets before training by preserving gradient flow.
\newblock {\em arXiv preprint arXiv:2002.07376}, 2020.

\bibitem{wang2012cost}
Tao Wang, Zhenxing Qin, Shichao Zhang, and Chengqi Zhang.
\newblock Cost-sensitive classification with inadequate labeled data.
\newblock {\em Information Systems}, 37(5):508--516, 2012.

\bibitem{Weiss2004MiningWR}
Gary~M. Weiss.
\newblock Mining with rarity: a unifying framework.
\newblock {\em SIGKDD}, 6:7--19, 2004.

\bibitem{Xia2021RobustEH}
Xiaobo Xia, Tongliang Liu, Bo Han, Chen Gong, Nannan Wang, ZongYuan Ge, and Yi
  Chang.
\newblock Robust early-learning: Hindering the memorization of noisy labels.
\newblock In {\em ICLR}, 2021.

\bibitem{yao2020searching}
Quanming Yao, Hansi Yang, Bo Han, Gang Niu, and James Tin-Yau Kwok.
\newblock Searching to exploit memorization effect in learning with noisy
  labels.
\newblock In {\em International Conference on Machine Learning}, pages
  10789--10798. PMLR, 2020.

\bibitem{zhang2021understanding}
Chiyuan Zhang, Samy Bengio, Moritz Hardt, Benjamin Recht, and Oriol Vinyals.
\newblock Understanding deep learning (still) requires rethinking
  generalization.
\newblock {\em Communications of the ACM}, 64(3):107--115, 2021.

\bibitem{zhang2010cost}
Shichao Zhang.
\newblock Cost-sensitive classification with respect to waiting cost.
\newblock {\em Knowledge-Based Systems}, 23(5):369--378, 2010.

\bibitem{zhang2012decision}
Shichao Zhang.
\newblock Decision tree classifiers sensitive to heterogeneous costs.
\newblock {\em Journal of Systems and Software}, 85(4):771--779, 2012.

\bibitem{zhang2021efficient}
Zhenyu Zhang, Xuxi Chen, Tianlong Chen, and Zhangyang Wang.
\newblock Efficient lottery ticket finding: Less data is more.
\newblock In {\em International Conference on Machine Learning}. PMLR, 2021.

\bibitem{zhou2019multi}
Steven Zhou, Yixin Zhuang, and Rusong Meng.
\newblock Multi-category skin lesion diagnosis using dermoscopy images and deep
  cnn ensembles.
\newblock {\em l{\'\i}nea], ISIC Chellange}, 2019.

\end{thebibliography}
}

\end{document}